\documentclass{llncs}

\usepackage{booktabs}
\usepackage{graphicx}
\usepackage{amsmath, amssymb, amstext}
\usepackage{multirow}
\usepackage{array}
\usepackage{bm}
\usepackage[caption=false,font=normalsize,labelfont=sf,textfont=sf]{subfig}

\newcommand{\specialcell}[2][c]{%
  \begin{tabular}[#1]{@{}c@{}}#2\end{tabular}}

\begin{document}
\title{Clinically Meaningful Comparisons Over Time: An Approach to
  Measuring Patient Similarity based on Subsequence Alignment}

\author{Dev Goyal, Zeeshan Syed, Jenna Wiens}
\institute{University of Michigan, Ann Arbor}
\maketitle

\begin{abstract}
  Longitudinal patient data has the potential to improve clinical risk
  stratification models for disease. However, chronic diseases that
  progress slowly over time are often heterogeneous in their clinical
  presentation. Patients may progress through disease stages at
  varying rates. This leads to pathophysiological misalignment over
  time, making it difficult to consistently compare patients in a
  clinically meaningful way. Furthermore, patients present clinically
  for the first time at different stages of disease. This eliminates
  the possibility of simply aligning patients based on their initial
  presentation. Finally, patient data may be sampled at different
  rates due to differences in schedules or missed visits. To address
  these challenges, we propose a robust measure of patient similarity
  based on \textit{subsequence alignment}. Compared to global
  alignment techniques that do not account for pathophysiological
  misalignment, focusing on the most relevant subsequences allows for
  an accurate measure of similarity between patients. We demonstrate
  the utility of our approach in settings where longitudinal data,
  while useful, are limited and lack a clear temporal alignment for
  comparison. Applied to the task of stratifying patients for risk of
  progression to probable Alzheimer's Disease, our approach
  outperforms models that use only snapshot data (AUROC of $0.839$
  vs. $0.812$)
  and models that use global alignment techniques (AUROC of $0.822$).
  Our results support the hypothesis that the \textbf{trajectories} of
  patients are useful for quantifying inter-patient similarities and
  that using subsequence matching and can help account for
  heterogeneity and misalignment in longitudinal data.
\end{abstract}

\section{Introduction}
\label{sec:introduction}

While the increasing availability of patient data holds out the
promise of better risk stratification models, many problems or
outcomes of interest are plagued with patient heterogeneity. That is,
a patient's trajectory through disease is regulated by complex
interactions that result from clinical, lifestyle, genetic and
environmental
factors~\cite{ginsburg2001personalized,hamburg2010path}. While such
trajectory information can help shed light on how patients progress
through disease, it can be difficult to make meaningful longitudinal
comparisons~\cite{alva2014effect}. In particular, at the time of
initial presentation, patients are often at varying stages of
disease. They may be grouped under coarse clinical labels that range
from early to late stage disease. Thus, simply aligning patients by
the time of enrollment may lead to inaccurate comparisons. Moreover,
disease may progress quickly or slowly depending on the patient. This
introduces pathophysiological misalignment leading to inconsistent
comparisons over time. Finally, patients may miss scheduled visits,
leading to disparity in the lengths of their temporal data and/or
sampling times.

In this work, we present an approach for patient risk stratification
that utilizes all available longitudinal patient data while addressing
the challenges mentioned above. We propose a measure of patient
similarity that compares longitudinal patient data using an
optimal-cost time-series matching algorithm based on dynamic time
warping (DTW)~\cite{rabiner1978considerations}. While DTW typically
assumes the beginning and end of time series to be aligned and
constricts their end-points to match, we relax this assumption. In
particular, we use a \emph{subsequence matching} approach, where the
ends of the time series need not be
aligned~\cite{muller2007dynamic}. This approach allows us to utilize
the most relevant longitudinal information while making fewer
assumptions about how to align patients in time. Furthermore,
subsequence matching is robust to variability in the lengths of time
series. Finally, the proposed subsequence matching approach
generalizes broadly since it does not require expert knowledge or an
extra hyperparameter to extract the most relevant parts of time
series~\cite{silva2016prefix}. Our main contributions are:

\begin{itemize}
\item we formulate the challenge of defining
  patient similarities based on longitudinal data as a minimum cost
  alignment problem,
\item we motivate and present an alignment
  method, based on subsequence matching, to compare longitudinal
  data from patients, and
\item we rigorously evaluate and demonstrate an improvement in
  predictive performance from emphasizing \emph{the most relevant
    data} using subsequence matching over other alignment techniques
  that ignore pathophysiology through their alignment constraints
  (such as global DTW, prefix and suffix matching)
\end{itemize}

We demonstrate the utility of our approach by applying it to the task
of predicting patient progression to probable Alzheimer's Disease
(AD), specifically progression from Mild Cognitive Impairment (MCI) to
probable AD. Aside from the challenges discussed above, predicting
progression to probable AD is particularly challenging because of the
poorly understood pathophysiology of the disease as well as the
variable clinical presentation of AD in patients. Applied to a
publicly available dataset of MCI patients, the subsequence matching
approach outperforms a global DTW approach that considers the entire
time series when calculating similarity between patients (AUROC of
$0.839$ vs. $0.822$).

The rest of the paper is organized as follows. In
Section~\ref{sec:related-work}, we briefly review related work in the
context of longitudinal data. In Section~\ref{sec:methods}, we
introduce notation and present our proposed similarity metric based on
subsequence matching. We present and discuss our experimental results
on real data in Sections~\ref{sec:exp-and-results}
and~\ref{sec:the-data}. Finally, in
Section~\ref{sec:summary-and-conclusions}, we summarize our
contributions and discuss its implications beyond this study.

\section{Related Work} 
\label{sec:related-work}

Time-series classification is a well-studied area of research. For an
in-depth review of sequence classification, we refer the reader
to~\cite{xing2010brief}. Briefly, most time-series classification
approaches focus on either (a) defining a measure of dis/similarity
between raw signals or (b) on extracting features such as
motifs/shapelets/statistical summaries from the
data~\cite{chiu2003probabilistic,ding2008querying,luo2016predicting,shokoohi2015discovery,wiens2012patient,syed2009relation}. We
focus on the first setting, in which we consider the entirety of the
signal. We believe this is particularly important in settings where
there is a paucity of data available. Such settings are common
important in the healthcare domain, where collecting patient data is
often expensive and labor intensive. In addition, we limit our
analysis to interpretable models. While others have proposed
time-series classification techniques using a deep learning
framework~\cite{razavian2016multi,thodoroff2016learning,lipton2016modeling},
such approaches do not apply to a setting like ours in which the
number of training points is relatively small and the time series
relatively short, and yield predictions that are hard to interpret
limiting their utility.

Existing approaches to time-series classification based on the raw
signals often assume there exists some mechanism for fiducial temporal
alignment. For example, in~\cite{wiens2012patient} Wiens et al.\ align
examples based on time of admission and in~\cite{syed2009relation}
Syed et al.\ align time series based on the phases of a heartbeat. In
contrast, we focus on a more general setting, in which we relax the
assumption that such an alignment mechanism is always available. We
are not the first to relax this assumption. In particular, Silva et
al.\ introduce prefix and sufix invariant dynamic time warping
(DTW)~\cite{silva2016prefix}, where the global DTW constraints are
relaxed up to a chosen tolerance parameter that is treated as a
hyperparameter. We build upon this idea allowing each pair of time
series to match subsequences that achieve the minimum cost for the
particular pair~\cite{muller2007dynamic}, thus obviating the need for
an extra hyperparameter. Furthermore, we assume that the entirety of
each time series is relevant. However, we allow for the notion of
relevance to vary across different pairs of time series. Thus, we can
accurately compare time series that are misaligned in time as well as
collected over different pathophysiological phases of the disease.

Time-series data are often multi-modal, \emph{i.e.},\ multiple
heterogeneous sources of data exist. While DTW is trivially applicable
to these settings, Shokoohi-Yekta et al.~\cite{shokoohi2015non} find
that the generalization of DTW to multi-modal time series is sensitive
to the distance metric used for multi-modal comparisons. Furthermore,
the application of DTW to multi-modal time series is further
complicated by block-wise missing in the data. For example, patients
in a study may receive an ubiquitous test (\emph{e.g.,} a blood draw)
on all visits but a specialized test (\emph{e.g.,} a lumbar puncture)
on only every other visit. While several approaches to deal with
missing data have been presented
before~\cite{van2014evaluating,xiang2013multi}, these approaches do
not consider multi-modal time series. In this paper, we specifically
deal with the case of block-wise missing data in multi-modal time
series.

\section{Methods}
\label{sec:methods}

We begin by introducing notation used throughout the
remainder of the paper. Next, we present the proposed subsequence
matching approach. This approach aligns patients longitudinally and
calculates a distance function $d\left(\bm{X}^i, \bm{X}^j\right)$
between a pair of time series.

\subsection{Notation}
\label{sec:notation}

We assume a dataset of $N$
patients, where each patient is associated with a sequence of
visits. A visit is associated with a feature vector and a label. Each
patient has a sequence of visits along with clinically assigned
labels,

\begin{align*}
  \{\bm{Y}^p,\bm{X}^p\}_{p=1}^N & \quad \\
  \text{where } \bm{X}^p &= \left[\bm{x}^p_1,\bm{x}^p_2,\ldots,\bm{x}^p_v,\ldots,\bm{x}^p_{V_p}\right],\\
  \bm{Y}^p &= \left[y^p_1, y^p_2, \ldots, y^p_v, \ldots, y^p_{V_p}\right]\\
  \bm{x}_{(\cdot)}^{(\cdot)} &\in \mathbb{R}^d,\quad \bm{X}^p \in \mathbb{R}^{d\times V_p}, \quad \bm{Y}^p \in \{0, 1\}^{V_p}
\end{align*}

where $p$
represents the patient index, $V_p$
is the number of visits of patient $p$,
$N$
is the number of patients and $d$
is the number of features. The feature vector $\bm{x}_v^p$
encodes the biomarker measures associated with patient $p$
at visit $v$,
and $y_v^p$
represents the label associated with that visit. In particular,
$y_v^p = 1$
represents a positive label (\emph{i.e.}, progression to disease as
discussed in Section~\ref{sec:the-data}).

For each patient $p$,
we aim to predict the probability of progression from MCI to probable
AD \emph{at each visit} starting at and including their third visit:

\begin{align*}
  \Pr(y^p_v &= 1|\bm{X}^p_{1:v}) \qquad \forall~p \in \left[1,\ldots,
              N\right] \\
  \bm{X}^p_{1:v} &= [\bm{x}^p_1,\bm{x}^p_2,\ldots,\bm{x}^p_v] \qquad \forall~v\in \left[3,\ldots,V_i\right]\\
\end{align*}

\begin{figure}[!t]
\centering
\includegraphics[width=0.9\textwidth]{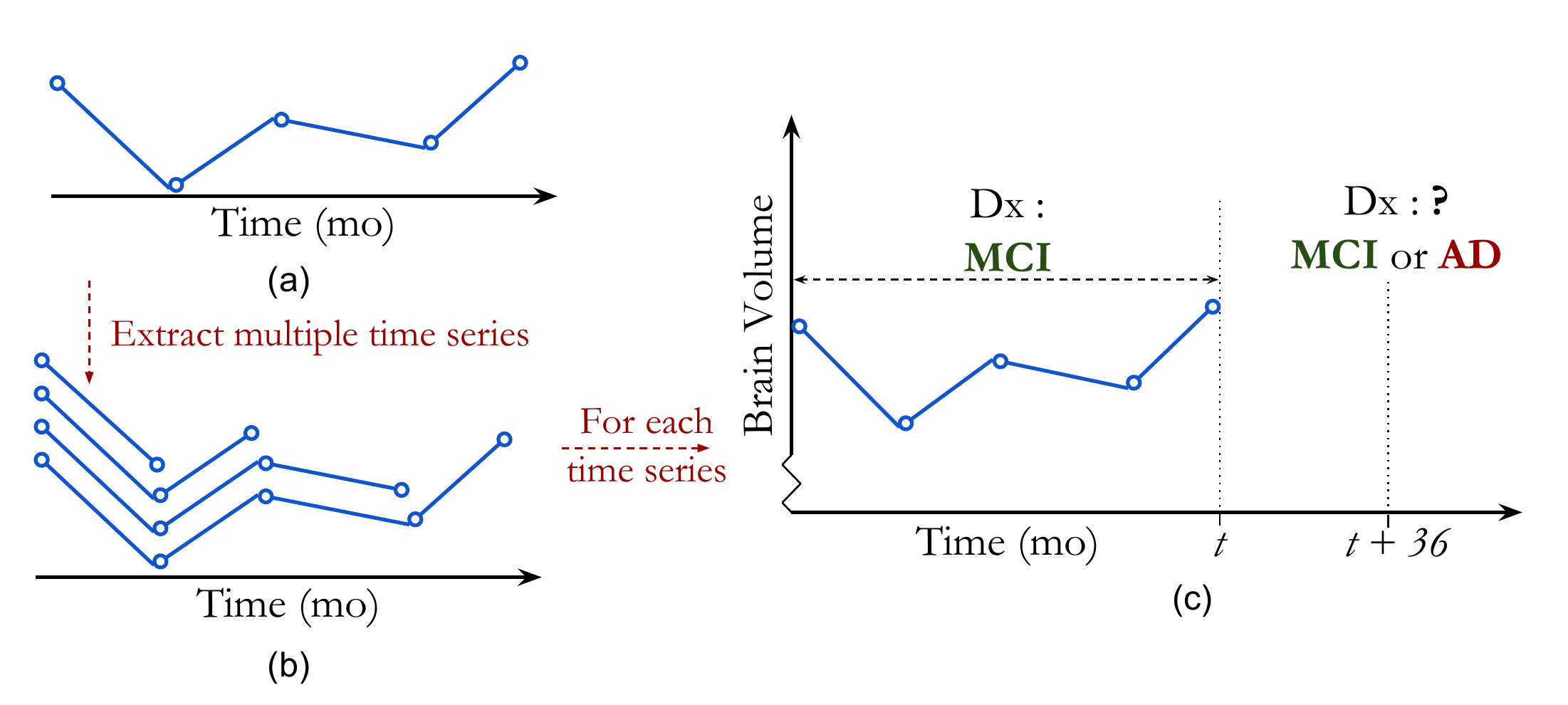}
\caption{Given longitudinal data for a patient (as in (a)), we (b)
  split the time series into several smaller time series. As shown in
  (c), we then predict progression from MCI to AD within a 36 month
  time horizon past the end of each time series.}
\label{fig:goal}
\end{figure}

It is worth noting that a prediction for a patient at a given visit is
based only on data from the past and present visits. Thus, a patient
with $V_p$
visits is represented by $\max(1, V_p-2)$
separate instances (each of which is a time series) in the dataset,
each with its own label (see Figure~\ref{fig:goal}). In what follows,
we use $\bm{X}^i$
to denote the $i^{th}$
data instance with length $V_i$,
with the understanding that several data instances may come from a
single patient. Thus, we have
$\bm{X}^i=\left[\bm{x}^i_1, \bm{x}^i_2, \ldots,\bm{x}^i_{V_i}\right]$
and $\bm{Y}^i=\left[y^i_1, y^i_2, \ldots,y^i_{V_i}\right]$.

\subsection{Subsequence Matching} 
\label{sec:subsequence-matching}

Given two time series $\bm{X}^i$
and $\bm{X}^j$,
we calculate a cost matrix $\bm{C} \in \mathbb{R}^{V_i\times V_j}$,
where $\bm{C}(v, w) = ||\bm{x}^i_v - \bm{x}^j_w||_2^2$.
This cost matrix is used to fill up an accumulated cost matrix
$\bm{D} \in \mathbb{R}^{(V_i+1, V_j+1)}$
in a recursive fashion as follows:

\begin{align*}
\bm{D}(1, w) &= 0 \qquad \forall~w\in \{1, \ldots, V_j\}\\
\bm{D}(v, 1) &= \infty \qquad \forall~v\in \{1, \ldots, V_i\}\\
\bm{D}(v, w) &= \bm{C}\left(v-1, w-1\right) + \min
\begin{cases}
\bm{D}(v-1, w),\\
\bm{D}(v, w-1), \\
\bm{D}(v-1, w-1) \\ 
\end{cases}\\
&\forall~v\in\{2, \ldots, V_i+1\}, w\in\{2, \ldots, V_j+1\}
\end{align*}

Following this, the distance between two time series is calculated as
$\min_w \bm{D}\left(V_i+1, w\right) \forall w\in\{2, V_j+1\}$.
Given time series $\bm{X}^i$
and $\bm{X}^j$,
we assume without loss of generality that $V_i \leq V_j$. When this is
not the case, the matrix $\bm{D}$ is transposed.

Compared to the traditional formulation of DTW that constrains the
ends of time series to match~\cite{rabiner1978considerations},
subsequence matching differs by allowing subsequences of the longer
time series to match the shorter time series~\cite{muller2007dynamic}
(i.e., not all data points from the longer time series are necessarily
included in the alignment). This is done to account for variability in
the lengths of the time series in our dataset as well as the variable
rates of disease progression that occur as a result of heterogeneity
among patients.

For the purpose of comparison, we also present results using prefix
and suffix matching~\cite{silva2016prefix}. In prefix matching, the
goal is to match some prefix $\bm{X}^j_{1:w}$
of time series $\bm{X}^j$
to $\bm{X}^i$
and achieve minimum cost alignment. Similarly, in suffix matching, the
goal is to match some suffix $\bm{X}^j_{w:V_j}$
of time series $\bm{X}^j$
to $\bm{X}^i$
and achieve minimum cost alignment. Compared to the formulation
in~\cite{silva2016prefix}, our prefix/suffix matching does not need an
extra hyperparameter to determine the relevant prefixes/suffixes in
the data. Instead, we use a minimum cost alignment approach to
determine the pre/suffix in a data-driven manner. Mathematically,
suffix matching is equivalent to prefix matching on an accumulated
cost matrix that is rotated $180$ degrees.

\section{The Data}
\label{sec:the-data}

To test the utility of our proposed approach, we consider a large
publicly available dataset pertaining to patients with AD.

\subsection{Study Population}
\label{sec:study-pop}

We use data made available by the Alzheimer's Disease Neuroimaging
Initiative (ADNI)~\cite{mueller2005alzheimer}. In ADNI, subjects with
MCI are recruited based on their cognitive test scores, with the goal
of enrolling patients in the early stages of MCI.\ Once enrolled,
patients are periodically examined over the duration of the study,
with frequencies of \mbox{6--24} months. At each examination, patients are
diagnosed as either MCI or probable AD on the basis of
questionnaire-based neuropsychological exams and the discretion of the
attending clinician. These clinical diagnoses serve as our ground
truth (typically, a gold standard diagnosis of AD required post-mortem
histopathological examination and is rarely available).

\begin{figure}[!t]
\centering
\includegraphics[width=3in]{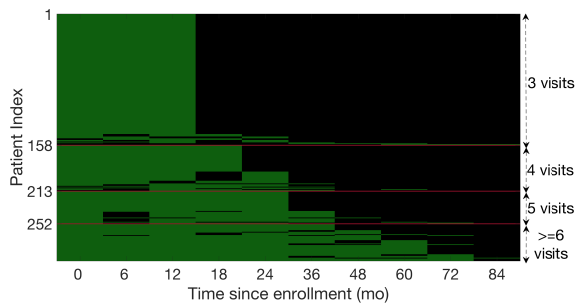}
\caption{Heat-map characterizing the missingness in our data. Rows
  represent the data available for a patient. Green cells represent
  available data for a patient at the particular time, whereas black
  cells represent missing data. The horizontal white lines divide the
  data into blocks with varying lengths. Even for patients that have
  the same number of visits, the actual schedule is often
  misaligned. This sporadic missingness makes it challenging to
  incorporate all longitudinal data by aligning patients
  chronologically.}
\label{fig:visit-map}
\end{figure}

While we focus on patients enrolled as MCI in this study, there is
still considerable heterogeneity among MCI patients in terms of the
extent of cognitive decline and their clinical
symptoms~\cite{ganguli2004mild}. Furthermore, patients progress to AD
at varying rates (between $3\mbox{--}8$
years) and have sporadic missingness in their visit schedules
(Figure~\ref{fig:visit-map} demonstrates the nature and extent of
missingness in our data).

In our study cohort, we use patients that have a clinical diagnosis of
MCI and also three or more visits (i.e., time points). Our final study
population consists of $511$
patients and $1264$
examples (since patients have multiple visits). For each visit with at
least 2 prior visits, we aim to predict whether or not a patient will
progress from MCI to probable AD within the next $36$
months (see Figure~\ref{fig:goal}). Among our $1264$
instances, $725$
(from 269 patients) remained stable as MCI whereas $539$
(from 258 patients) progressed to probable AD within $36$
months. Note that a single patient can contribute both positive and
negative instances to the dataset.

\subsection{Features}

At each patient visit ADNI collects a variety of data including, but
not limited to, MRI scans, PET scans, neuropsychological scores,
etc. We focused on MRI and FDG-PET data since almost all patients in
ADNI have an MRI scan ($\approx 35\%$
have a PET scan) performed on them at every visit. Moreover, several
studies have identified brain volume and metabolism as an important
biomarker of AD~\cite{weiner2013alzheimer,frisoni2010clinical}. We use
brain volumes and glucose uptake in patients extracted from MRI and
PET scans as features, and represent the features collected over
multiple visits as a multivariate time series. The raw MRI and PET
scans were processed using Freesurfer and these data were made
publicly available by the ADNI MRI team~\cite{jack2008alzheimer}.


\section{Experiments and Results}
\label{sec:exp-and-results}

To evaluate the utility of our proposed approach on the task
described above, we compare to a number of different approaches. In
this section, we begin by describing these approaches, then move on to
our experimental setup and finally results.

\subsection{Comparison Methods}
\label{sec:comparison-methods}

In Section~\ref{sec:methods}, we proposed subsequence matching as a
measure for inter-subject comparisons based on longitudinal data. In
this section, we compare the discriminative power of the proposed
approach to the following methods for measuring patient
similarity/differences.

\begin{description}
\item[Snapshot] The baseline approach that uses only the most recent
  patient visit for comparison, thus ignoring temporal data.
\item[Global DTW] We also consider a standard application of DTW,
  where the time series are constrained to match in entirety from
  beginning to end~\cite{rabiner1978considerations}.
\item[Prefix Matching] Compared to global DTW, prefix
  matching~\cite{silva2016prefix} constrains only the beginning of the
  two time series to match and is thus more constrained than
  subsequence matching.
\item[Suffix Matching] Finally, for completeness we consider a
  modification of the prefix matching approach, where only the ends of
  the time series are constrained to match.
\end{description}

\subsection{Evaluation}
\label{sec:evaluation}

In order to evaluate the performance of the proposed subsequence
matching approach, we apply the inter-patient similarities calculated
by it to the task of predicting progression from MCI to probable AD
within $36$
months (see Figure~\ref{fig:goal}). In particular, we use the
pair-wise distances between time series as features for the prediction
task. When using snapshot data, we use the biomarkers at the most
recent visit as features.

\subsection{Experimental Setup}
\label{sec:exp-setup}

In the following sections, we present results to test the following
hypotheses:
\begin{itemize}
\item incorporating longitudinal data improves predictive performance
  compared to using snapshot data only,
\item using relevant subsequences of longitudinal data to compare
  patients is more accurate than comparing entire time series
\end{itemize}

We test the second hypothesis by comparing the subsequence matching
approach that emphasizes the most relevant data with the global DTW
approach that constrains the time series to match in their entirety.

For each patient, the pair-wise distance of their time series with
every other patient, using both MRI and PET features, in the training
data serves as the feature vector for the classification model. For
patients that did not have PET scans, we used explicit matrix
factorization~\cite{hu2008collaborative} to impute the missing
distance measures.

We used L2-regularized logistic regression as our classifier,
implemented using the LIBLINEAR~\cite{liblinear} package. We perform
leave-one-patient-out testing, where all data belonging to a single
patient are left out in a particular test fold. All hyper-parameters
were chosen through a nested cross-validation performed on the
training data alone. We used the area under the ROC curve (AUROC)
metric to evaluate our classifiers. We use the method presented by
DeLong et al.~\cite{delong1988comparing} to compute $95\%$
confidence intervals and to perform statistical significance tests to
compare competing prediction methods (significance level was set at
$5\%$). All reported $p$ values are based on a two-sided z-test.

We note that a potential issue with this dataset could be the bias
introduced by longer time series. In particular, longer time series
could be more likely to eventually test positive and end up biasing
the classifier as a result. However, this is not an issue because we
make multiple predictions for each time series (see
Figure~\ref{fig:goal}), thus patients with long time series who
eventually test positive are also represented by short time
series. Thus, any potential correlation between the lengths of time
series and their labels is eliminated.

\subsection{Results and Discussion}
\label{sec:results-and-discussion}

The results of our experiments are given in Table~\ref{table:results}
and discussed below. Compared to using snapshot data only, using
longitudinal data that is aligned by using any of the DTW based
methods leads to an improvement in performance. The subsequence
matching approach outperforms the other DTW based methods of prefix
matching ($p<0.01$),
suffix matching ($p<0.01$) and global DTW ($p<0.01$).

\begin{table*}[!t]
  \renewcommand{\arraystretch}{1.1}
  \caption{Comparing predictions from using snapshot data only (Row 1)
    with the various approaches that incorporate longitudinal data
    (partial and complete). Subsequence matching based on DTW performs 
    best among all approaches.}
\label{table:results}
\centering
\begin{tabular}{c|c||c|c}
  \textit{Data} & \textit{Alignment Method} & \textit{AUROC ($95\%$ CI)}\\
  \hline
  Snapshot & Not Applicable & $0.812$ ($0.785 \mbox{--} 0.840$)\\
  \hline
  \multirow{4}{*}{\specialcell{Longitudinal}} 
                & Prefix Matching & $0.813$ ($0.786 \mbox{--} 0.841$)\\
                & Suffix Matching & $0.820$ ($0.793 \mbox{--} 0.847$)\\
                & Global DTW & $0.822$ ($0.795 \mbox{--} 0.849$)\\
                & \textbf{Subsequence Matching} & $0.839$ ($0.814 \mbox{--} 0.865$)\\
\end{tabular}
\end{table*}

\subsubsection{Longitudinal vs Snapshot models}
\label{sec:long-vs-snapshot}

\begin{figure}[!t]
\centering
\subfloat[]{\includegraphics[width=0.4\textwidth]{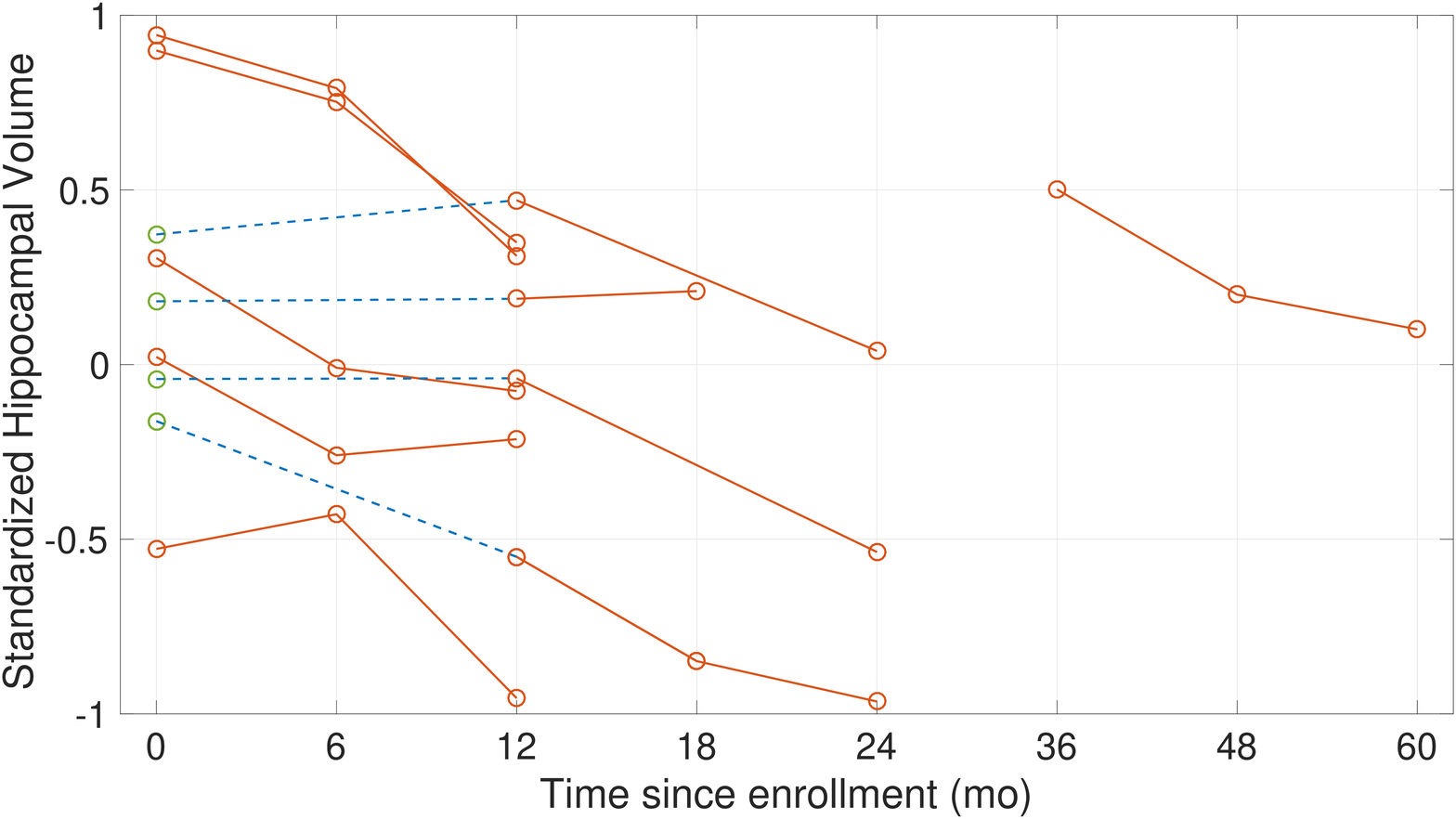}%
\label{fig:sub-vs-snap}}
\hfil
\subfloat[]{\includegraphics[width=0.4\textwidth]{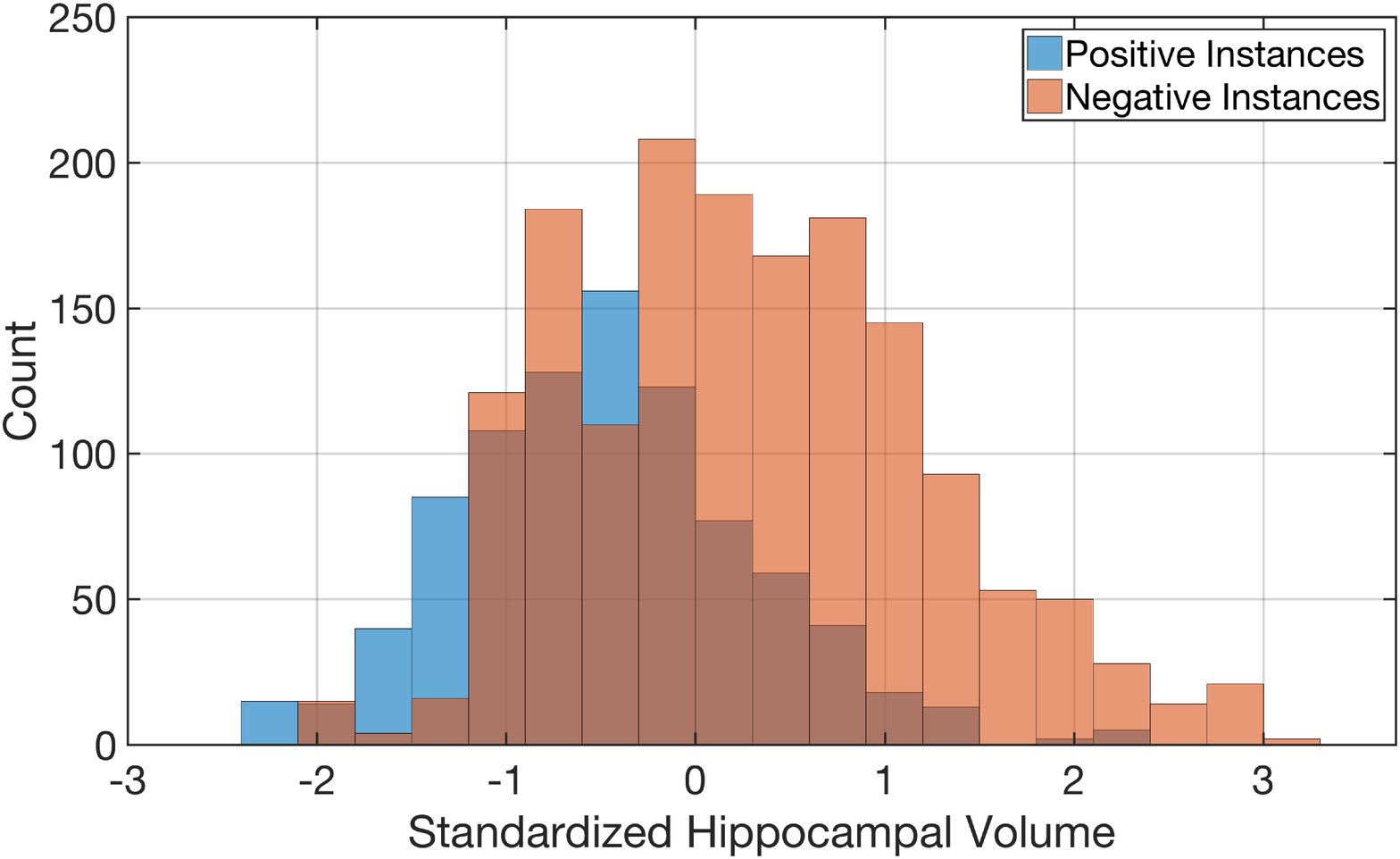}%
\label{fig:hippocampal-spread}}
\caption{(a) Representative time series that progress to AD where
  subsequence matching outperformed a model based on snapshot data
  only. Red markers represent visits where the patient had a positive
  label (will progress to AD within 36 months), green markers
  represent a negative label and the dotted blue lines represent the
  period during which the label went from negative to positive. These
  instances are characterized by a pronounced decline in brain volume
  despite the ultimate snapshot value being close to or above the
  average value for at least some of them. (b) Comparing the
  distributions of the hippocampal volume for positive and negative
  instances. These distributions have considerable overlap, making it
  difficult to classify them using snapshot values alone.}
\end{figure}

To understand the source of the large improvement in performance from
using subsequence matching compared to snapshot features, we
constructed contingency tables to discover the examples where
subsequence matching outperformed snapshot features (we used a cutoff
of $0.5$
to classify a patient as positive). The main source of improvement was
$83$
positive instances (that came from $53$
unique patients), where subsequence matching predicted correctly and
snapshot did not ($p < 0.01$).
As shown in Figure~\ref{fig:sub-vs-snap}, these instances are
characterized by a pronounced decline in brain volume close to the
progression from MCI to AD.\ In comparison, the ultimate hippocampal
volumes of these instances (i.e.\ the features for the snapshot model)
show considerable overlap for the positive and negative instances (see
Figure~\ref{fig:hippocampal-spread}). We chose to visualize the
hippocampal volume as it receives the highest weight in the snapshot
models, and is well known to be an important predictor of
AD~\cite{jack2010hypothetical}. We visualize the top $10$
instances that have the highest differences in predicted probability
(these differences were at least $50\%$)
as predicted by the two approaches. This suggests that a decline in
brain volume is an important predictor of disease progression, more so
than low brain volume alone.

\subsubsection{Subsequence Matching vs Global DTW}
\label{sec:subsequence-vs-global}

The overall AUROC of global DTW was 0.822. Subsequence matching
outperformed global DTW with an AUROC of 0.839 ($p<0.01$).
Our results suggest that the instances where subsequence matching
outperformed global DTW were characterized by longer time
series. Among the instances where subsequence matching outperformed
global DTW, the time series were nearly twice as likely ($35\%$
vs $17\%$
probability) to have 5 or more visits compared to the overall
data. Intuitively, we believe the source of this improvement is the
exponential distribution of the number of visits the time series in
our data have. In particular, given that the vast majority of our data
have 3 or fewer visits ($68\%$),
allowing the longer time series to match partially with these shorter
time series allows for a more accurate measure of similarity between
them.

\section{Summary and Conclusions}
\label{sec:summary-and-conclusions}

In contrast to existing methodologies that use snapshot or
cross-sectional data to stratify patients by risk of progression to
disease, in this study we explored incorporating all available
longitudinal patient data. While approaches exist for leveraging
longitudinal data, they often assume the availability of some fiducial
marker for temporal alignment. In contrast, we propose and evaluate an
approach for comparing variable length patient time series that lack
such a fiducial marker. We consider a measure of similarity based on
minimal cost alignment subsequence matching. Our approach accounts for
heterogeneous rates of decline in patients by non-linearly warping the
data during the alignment process, while focusing on the most relevant
data.

We demonstrate the utility of our proposed similarity measure on the
task of predicting which patients at an intermediate disease stage
(MCI) are most likely to progress to AD within 36 months. The propose
similarity measure applies despite the variability in the lengths of
the time series. In the ADNI dataset, the median number of visits per
patient is $3$,
but this ranges from $2$
to $9$,
with about $32\%$
of patients have $4$
or more visits. Applied to these data, the proposed approach achieved
an AUROC of $0.839$,
outperforming other non-linear alignment techniques.

While we focused on the challenging task of predicting progression to
AD, the proposed approach for measuring patient similarity based on
longitudinal data could apply more broadly. In particular, this
technique is applicable to other settings that lack a meaningful
fiducial marker for alignment and in which disease progression
manifests itself variably across patients.


\end{document}